\documentclass{article}

\usepackage{microtype}
\usepackage{graphicx}
\usepackage{caption}
\usepackage{subfigure}
\usepackage{booktabs} %
\usepackage{amssymb}
\usepackage{color, colortbl}
\usepackage{footnote}

\usepackage[table, svgnames]{xcolor}

\usepackage{tikz}
\def\checkmark{\tikz\fill[scale=0.4](0,.35) -- (.25,0) -- (1,.7) -- (.25,.15) -- cycle;}
\def\abs{\@ifstar{\oldabs}{\oldabs*}}

\usepackage{amsmath}

\usepackage{hyperref}

\usepackage{xcolor}
\definecolor{darkgreen}{rgb}{0,0.6,0}
\definecolor{darkred}{rgb}{0.7,0.0,0}
\definecolor{darkblue}{rgb}{0,0.0,0.6}
\definecolor{magenta}{rgb}{0.8,0.1,0.8}
\definecolor{darksomething}{rgb}{0,0.4,0.6}

\usepackage[accepted]{icml2021}

\icmltitlerunning{Learn2Hop: Learned Optimization on Rough Landscapes}

\newif\ifcomments
\commentsfalse

\ifcomments
     
    \newcommand{\acom}[1]{{\color{purple}AM: #1}} 
    
\else
    \newcommand{\acom}[1]{{}} 
    
\fi

\begin{document}

\twocolumn[
\icmltitle{Learn2Hop: Learned Optimization on Rough Landscapes \\ With Applications to Atomic Structural Optimization}
\icmlsetsymbol{equal}{*}
\begin{icmlauthorlist}
\icmlauthor{Amil Merchant}{goo,resident}
\icmlauthor{Luke Metz}{goo}
\icmlauthor{Sam Schoenholz}{goo}
\icmlauthor{Ekin Dogus Cubuk}{goo}
\end{icmlauthorlist}
\icmlaffiliation{goo}{Google Research, Mountain View, California, USA}
\icmlaffiliation{resident}{This work was done as part of the Google AI Residency Program (https://research.google/careers/ai-residency/)}

\icmlcorrespondingauthor{Amil Merchant}{amilmerchant@google.com}
\icmlcorrespondingauthor{Ekin D. Cubuk}{cubuk@google.com}

\icmlkeywords{Machine Learning, Optimization, Learned Optimizers, Physics, Structured Prediction}
\vskip 0.3in ]
\printAffiliationsAndNotice{}
\begin{abstract}
Optimization of non-convex loss surfaces containing many local minima remains a critical problem in a variety of domains, including operations research, informatics, and material design. Yet, current techniques either require extremely high iteration counts or a large number of random restarts for good performance. In this work, we propose adapting recent developments in meta-learning to these many-minima 
problems by \textit{learning} the optimization algorithm for various loss landscapes. We focus on problems from atomic structural optimization---finding low energy configurations of many-atom systems---including widely studied models such as bimetallic clusters and disordered silicon.
We find that our optimizer learns a `hopping' behavior which enables efficient exploration and improves the rate of low energy minima discovery. 
Finally, our \textit{learned optimizers} show promising generalization with efficiency gains
on never before seen tasks (e.g. new elements or compositions). Code will be made available shortly.
\end{abstract}

\section{Introduction}
Efficient global optimization remains a problem of general research interest, with applications to a range of fields including operations design \cite{Ryoo_Global},
network analysis \cite{Abebe_Application}, and bio-informatics \cite{Liwo_Protein}. Within the fields of chemical physics and material design, efficient global optimization is particularly important for finding low potential energy configurations of isolated groups of atoms (clusters) and periodic systems (crystals); identifying low energy minima in these cases can yield new stable structures to be experimentally produced and tested for a wide variety of industrial or scientific applications  \cite{Wales_Global, Flikkema_Dedicated}. However, even simple examples with a few atoms have quite complex minima structures. Numeric approximations suggest that systems of only 147 atoms could have $10^{60}$ distinct minima \cite{Tsai_Use}. 

\begin{figure}[t]
    \centering
    \includegraphics[width=0.40\textwidth]{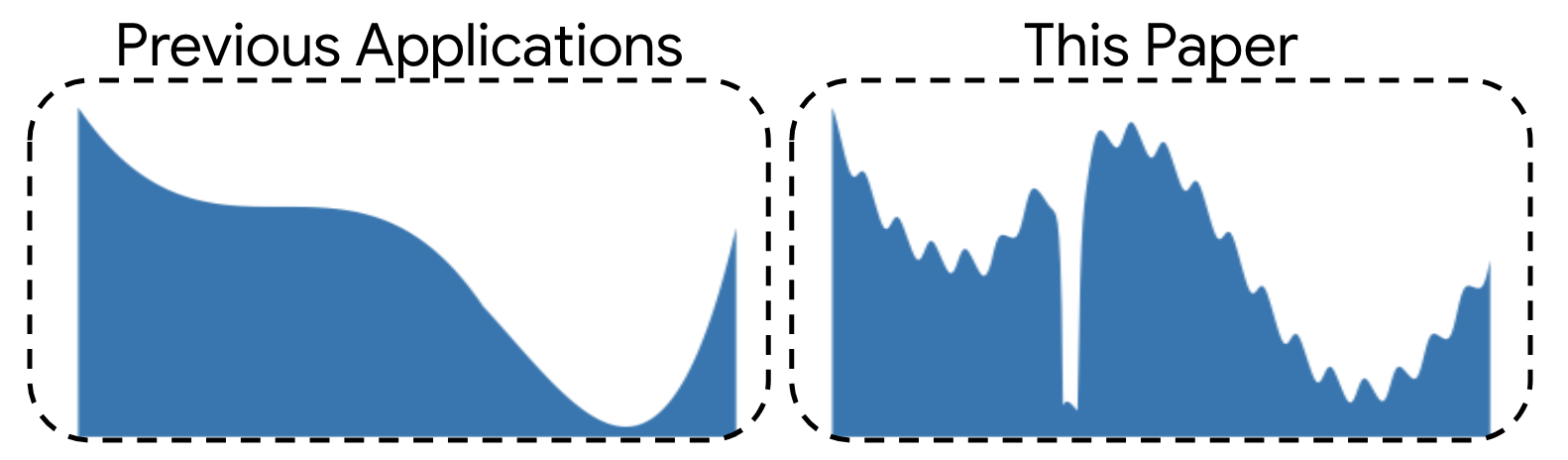}
    \vspace{-0.5em}
    \caption{Schematic diagram of the difficulties of global optimization on rough loss landscapes. In contrast to prior work where loss surfaces are approximately convex, this paper focuses on global optimization problems where minima are numerous and there is unlikely to be a low loss path between local minima.}
    \label{fig:schematic}
    \vspace{-1.0em}
\end{figure}

Global optimization problems can also be quite difficult when high loss barriers exist between local minima, as depicted in Figure \ref{fig:schematic}.\footnote{See \citet{Wales_Global} for examples of difficulties in optimizing Lennard-Jones potentials.} Despite being NP-hard in the worst case, significant work has been put into developing optimization techniques for these structure prediction tasks. Nonetheless, classical approaches to this problem continue to face a number of drawbacks including requirements of: a significant number of steps \cite{Wales_Global, Pickard_Ab}, carefully selected hand-crafted features, or sensitive dependence on learning rate schedules \cite{Bitzek_Structural}.

\begin{figure*}[t]
    \centering
    \includegraphics[width=0.7\textwidth]{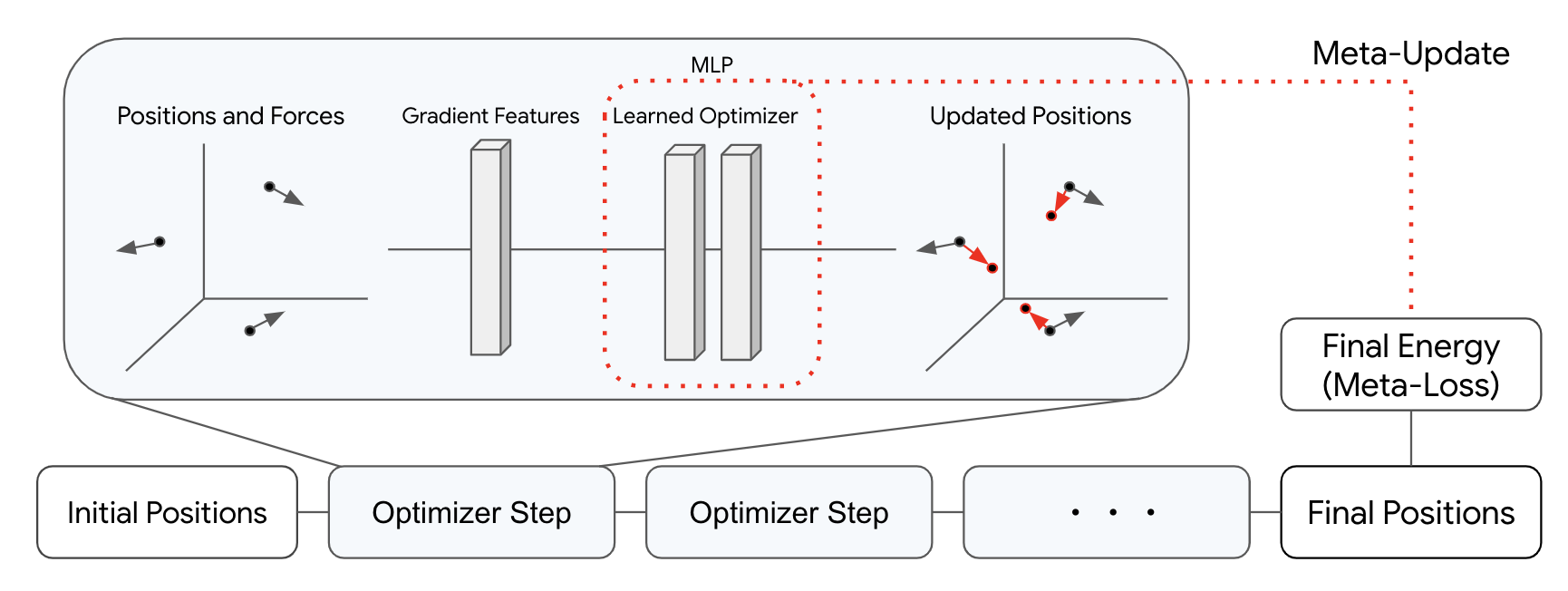}
    \vspace{-0.5em}
    \caption{Schematic diagram of a learned optimizer for atomic structural optimization. Positions and forces are computed from physical simulations (i.e. empirical potentials). Gradients are accumulated, featurized, and inputted to a shallow neural network that updates positions. Although the diagram specifies a problem in $\mathcal{R}^3$, the learned optimizer framework can be applied to arbitrary global optimization problem.}
    \label{fig:example}
    \vspace{-0.5em}
\end{figure*}

In this work, we propose adopting a new class of strategies to these global optimization problems: \textit{learned optimization} \citep{bengio1992optimization, Andrychowicz_Learning, Metz_Learned}. Here, hand-designed update equations are replaced with a learned function parameterized by a neural network and trained via meta-optimization. While this strategy has shown promise for training neural networks \cite{metz2020tasks} where falling into bad local minima is not a concern \cite{choromanska2015loss, Lee_No}, current techniques fail to prioritize global minimum discovery or have not been tested on rough loss landscapes with many unconnected local minima.

In this paper, we show that learned optimizers can offer a substantial improvement over classical algorithms for these sorts of global optimization problems. To this end, we present several modifications of learned optimizers required to effectively train models which can find low energy states of many-minima loss surfaces. Using several canonical problems in atomic structural optimization, we demonstrate that the learned optimizers outperform their classical counterparts when trained and tested on similar systems and---more surprisingly---are able to generalize to unseen systems. The specific contributions of this paper are as follows:

\begin{enumerate}
    \itemsep0em
    \item Novel parameterizations of learned optimizers prioritize global minimum discovery (Section \ref{sec:methodology}) and yield improvements on benchmark tasks consisting of single atom types (Section \ref{sec:single}).
    \item Analysis of learned optimizer behavior showcases an automatically-learned `hopping' behavior which enables efficient exploration of minima (Section \ref{sec:analysis}).
    \item Results for bimetallic systems show that our learned optimizers can generalize beyond the examples seen during training, yielding gains in efficiency and performance over commonplace optimization techniques such as Basin Hopping (Section \ref{sec:bimetallic}).
\end{enumerate}

\section{Background / Related Work}
\subsection{Atomic Empirical Potentials}
\label{sec:potentials}
Atomic structure optimization often requires finding the lowest energy configuration of a system \citep{oganov2019structure}. However, accurate calculation of energies is expensive, often requiring quantum mechanical simulations such as DFT \citep{hohenberg1964inhomogeneous}. In this work, we instead use approximations of the potential energy known as empirical potentials, that are not only simple and efficient to calculate but also have minima that correlate to those found by more accurate calculations \cite{Tran_Study}. The empirical potentials studied in this paper are as follows: 

\textbf{Lennard-Jones Clusters} are often used in the modelling of spherically-symmetric particles in free-space such as noble gasses or methane and are the archetypal model for a simple-to-compute potential \cite{Jones, Doye_Evolution}. The total energy of the system is defined only by pairwise distances, denoted $d_{ij}$:
\begin{equation}
    \sum_i \sum_{j>i} \epsilon \left[ \left( \frac{d_{0}}{d_{ij}} \right)^{12} - 2 \left( \frac{d_{0}}{d_{ij}} \right)^6 \right]
\end{equation}

where $\epsilon$ is the minimum two-particle energy and $d_{0}$ is the distance where this occurs. Despite the apparent simplicity, the minima structures of these systems are complex and vary significantly based on the number of atoms \cite{Doye_Evolution, Wales_Global}.\footnote{Diagrams depicting the local minima structures and the lowest energy paths between minima for particularly interesting cluster sizes can be viewed at \scriptsize{\url{http://doye.chem.ox.ac.uk/research/forest/LJ.html}}}
For example, the 13 and 19 atom systems display concentrated ``funnels'', where many local minima and the global minimum are connected via low energy paths. In contrast, the 38 and 75 atom counts display complex ``double-funneled'' landscapes, where there are two distinct sets of minima that are dynamically inaccessible due to a high energy barrier between the two.

\textbf{Gupta Clusters} are similar to the Lennard-Jones model in approximating the energy of sets of atoms in free-space, yet they are significantly more complex due to the inclusion of a second-moment approximation of the tight-binding Hamiltonian \cite{Gupta, Au, sutton1993electronic, Rosato}. In this paper, we focus on both single element and bimetallic forms using the elements Ag, Au, Pd, and Pt. Energy equations and the constant values for all systems are provided in Appendix \ref{app:gupta}.

\textbf{Stillinger-Weber (SW)} potentials \citep{stillinger1985computer} provide more accurate estimations for energies of semiconductors. This empirical potential introduces a three-body angular term between atoms, making the corresponding loss landscape significantly more difficult to optimize. In this paper, the SW potential is used to model silicon crystals. This benchmark is distinct from the others in the use of periodic boundary conditions, so that atomic structures are tiled in space. The associated energy equation and parameters are provided in Appendix \ref{app:sw}.

\subsection{Optimization Methods from Structure Prediction}
Early approaches to structure prediction problems simply initialized the particle positions at hand-crafted, physically-motivated structures, before applying gradient descent \cite{Hoare_Physical, Farges_Cluster, Doye_Effect}. This technique proved effective for simple cluster systems such as Lennard-Jones but faced difficulty scaling to more complex potentials (such as those with angular dynamics). Classic optimization approaches such as Basin Hopping \cite{Wales_Global} and Simulated Annealing \cite{Kirkpatrick_Optimization, Biswas_Simulated} resulted in significant improvements and helped discover the minima for a variety of structures. However, these techniques end up requiring high step counts and may only find the global minimum in the limit of infinite optimization steps.

Modern molecular dynamics systems use a variety of techniques for optimization. Quasi-Newton techniques such as BFGS and damped Beeman dynamics \cite{Beeman_Some} are popular within libraries such as QuantumEspresso \cite{Giannozzi_Quantum}. Alternate strategies include Fast Inertial Relaxation Engine--referred to as FIRE \cite{Bitzek_Structural}-- which adaptively modifies the velocity over the course of training and Ab Initio Random Structure Search \cite{Pickard_High, Pickard_Ab, Zilka_Ab}. However, these techniques often rely on heuristics or require a large number of restarts before reaching the global minimum.

While this work only uses traditional empirical potentials, machine learning has also been used to create empirical potentials, such as those utilizing graph convolutions \cite{gilmer2017neural,schutt2017schnet,cheon2020crystal}. These models are becoming a popular option for speeding up optimization. However, we note that the approach presented in this paper is complementary; the two could be combined so that both the potential and optimizer are learned. 

\subsection{Learned Optimization}
\label{sec:background}
Learned optimization \cite{bengio1992optimization, Andrychowicz_Learning, Wichrowska_Learned, lv2017learning, Metz_Learned, Metz_Using, gu2019meta, metz2020tasks}, has recently become a popular meta-optimization task, where updates are a function of the gradients, parameterized by a neural network. In the traditional setup depicted in Figure \ref{fig:example}, training a learned optimizer consists of an inner-loop of optimization problems which are used to compute meta-updates to the learned optimizer parameters, referred to as the outer-loop \cite{Wichrowska_Learned, Metz_Learned} .

In our case, the inner-loop consists of instantiations of atomic structure optimization problems, including a random initialization for atoms and a corresponding empirical potential to minimize. At each step in the inner-loop of meta-training, atomic forces are computed, featurized, and then input to the learned optimizer which computes updates to the particles. These steps are then repeated, which is often referred to as an inner-trajectory or unroll.

For each inner-loop, a meta-loss is defined based on the optimization trajectory, commonly the average loss over the trajectory in prior work. If the unrolls were short, meta-training could be performed by gradient descent \cite{Andrychowicz_Learning, Wichrowska_Learned}. However, due to memory requirements and often ill-conditioned outer-loss surfaces \citep{Metz_Understanding}, meta-gradients are instead approximated via Evolutionary Strategies (ES) using antithetic samples \cite{Williams_Simple, salimans2017evolution, Metz_Understanding}. A central controller collects batches of meta-gradient estimates and updates the learned optimizer parameters, typically using Adam \cite{Kingma_Adam}.

A variety of architectures have been proposed for learned optimizers. Early work utilized RNNs in order to provide the network a state that can be automatically updated throughout the course of training \cite{Andrychowicz_Learning}. These models were quickly developed into optimizers that scale \cite{Wichrowska_Learned}, and are compute-efficient \cite{Metz_Learned}. Most closely related to our work is that of \cite{Metz_Understanding, Metz_Learned} which is novel in its paramerization of the state and input features of the learned optimizer. Instead of providing an explicit memory (e.g. in a GRU), the learned optimizer is simplified to an MLP that is applied per parameter and is provided relevant features, such as the first and second moment estimates for the gradients. See Table \ref{tab:features} for all features used in the MLPs.

\section{Modifications for Rough Landscapes}
\label{sec:methodology}
Adapting these learned optimizers to many-minima landscapes requires modifications to both the training and model itself to improve global optimization. For example, instead of the average loss of the optimization trajectory, the learned optimizer for atomic structure optimization only uses the final step loss. This training strategy prioritizes global minimum discovery at the expense of greater variance of the gradient with respect to learned optimizer weights. Additional modifications are detailed in the following sections and training details are in Appendix \ref{app:comparison}.

\subsection{Features}
\label{sec:features}
We follow prior work \cite{Metz_Understanding} in parameterizing the learned optimizer as an MLP. The input features are often inspired by popular optimization techniques and include estimates of the first and second moments to mimic Adam \cite{Kingma_Adam}. Table \ref{tab:features} describes all inputted features that are adopted from ``MLPOpt'', the learned optimizer described in \citet{Metz_Understanding}.

\begin{table}[tb]
\caption{Features inputted into the learned optimizers. MLPOpt refers to the model by \citet{Metz_Understanding}.}
\vspace{-0.5em}
\label{tab:features}
\begin{center}
\begin{small}
\begin{sc}
\addtolength{\tabcolsep}{-4pt}
\begin{tabular}{lrr}
\toprule
Features & Description & In MLPOpt \\
\midrule
Gradients & Gradients & \checkmark \\ 
Positions & Particle positions & \checkmark \\
Decays & EMA of 1st and 2nd moments & \checkmark \\
Adam-Like & \shortstack[r]{Inverse norm and \\ moment correction} & \checkmark \\ 
Singular & Number of particles & \checkmark \\
Species & Species identity & \checkmark \\
Step & Training step sine features & \checkmark \\
\midrule
Radial & Radial symmetry features \\
\bottomrule
\end{tabular}
\addtolength{\tabcolsep}{6pt}
\end{sc}
\end{small}
\end{center}
\vspace{-1.5em}
\end{table}

Novel to our learned optimizers is the inclusion of Behler-Parrinello radial symmetry features \cite{Behler_Generalized, Artrith_Neural,cubuk2015identifying}. Traditional learned optimizers update each parameter independently, yet in the case of atomic structure problems, particle behavior should depend on interactions with nearest neighbors. Radial symmetry functions provide these sorts of two-body interactions for a central atom by allowing updates to be defined by local neighborhoods. Simply put, these features $\phi$ are computed using a Gaussian kernel and summing over all neighbors of a central atom. Smooth cutoffs $\Gamma$ are applied using the formulation by \citet{Behler_Generalized}:
\begin{align*}
    \phi_i &= \sum_{j \neq i} \exp \left( - \eta d_{ij}^2 \right) \Gamma(d_{ij}) \\
    \Gamma(d_{ij}) &=     \begin{cases}
        0 \text{ if } d_{ij} > c \\
        0.5 \left( \cos \left(\pi \cdot d_{ij} / c \right) + 1 \right) \text {otherwise}
    \end{cases}
\end{align*}
 where $c$ is a pre-defined cutoff set to 2.5 angstroms and $\eta$ is a hyperparameter controlling the scale. $\eta$ is set to one of $\{0.0009,0.01, 0.02, 0.035, 0.06, 0.1, 0.2, 0.4\}$, yielding 8 radial features per atom type.

These features are then parameterized into a log-magnitude and direction representation  \cite{Andrychowicz_Learning}:
\begin{equation*}
    \label{eq:parameterization}
    \text{features } = \begin{cases}
    \left( \log \lvert x \rvert  / p, \text{sign}(x) \right) &\text{ if } x > \exp(-p) \\
    \left( -1, x \exp(p) \right)  &\text{ if } x \leq \exp(-p)
    \end{cases}
\end{equation*}
where $p=10$ is the default hyperparameter. These features are then input to the learned optimizer, a small 2-layer dense neural network (with a hidden size of 32), that acts component-wise. The network outputs magnitude $m$ and unnormalized direction $d$ per component, converted to the final update via: 
\begin{equation}
    \alpha \cdot d \cdot \text{sigmoid}(\beta \cdot m + \gamma)
\end{equation}
where $\alpha, \beta, \gamma$ are scalars learned via meta-optimization.\footnote{Note, this output parameterization contrasts from \citet{Metz_Understanding}, but experimental evidence showed that the traditional exponential update leads to divergent optimization trajectories.} 

\subsection{Meta-Training Stability}
The rough loss landscapes discussed in this paper present two significant challenges with regards to meta-optimization: high curvature and infrequent training signal.

High curvature is an intrinsic problem to atomic structures. For example, with the Lennard-Jones potentials, the energy increases at a rate of $d_{ij}^{-12}$ as $d_{ij} \rightarrow 0$ for all $i,j$. When the optimizer happens to bring two particles too close together, energy (loss) spikes can yield meta-gradients that destabilize learned optimizer training. Traditional strategies such as gradient norm clipping \cite{pascanu2013difficulty} were found to be ineffective in preventing divergence of the meta-optimization objective. 

Infrequent and noisy training signals are also problematic as learned optimizers can find simple, stable optimization strategies such as gradient descent early in training. Most perturbations to gradient-descent methods will be noise and increase the final loss. The meta-optimization model must be sensitive enough to learn from the infrequent signal occurring when few individual instantiations of a learned optimizer find better minimia structure, rather than being pushed back to descent-like methods due to noise.

As mentioned in the background, many learned optimizers are trained with antithetic ES sampling \cite{salimans2017evolution, Metz_Understanding}
where meta-gradients are estimated via perturbations of the parameters: 
\begin{equation}
    \nabla_{meta} = \mathbb{E}_{\epsilon \sim \mathcal{N}(0, I\sigma^2)} \left[\frac{L(\theta + \epsilon) - L(\theta - \epsilon)}{2\sigma^2} \right] \epsilon
\end{equation}

where $L$ is the loss, $\theta$ the learned optimizer parameters, and
$\sigma$ is the perturbation scale, set to $0.1$. This strategy is particularly vulnerable to the optimization difficulties, as either direction of the parameter perturbation may lead to exploding gradients. Also, the variance of the estimator makes it more difficult to learn from the sparse rewards when optimizers find better local minima. To overcome these issues, we present two modifications to the meta-update which enable stable meta-training:

\paragraph{Meta-loss Clipping (ESMC)} \mbox{}\\
In order to prioritize signal from perturbations that find better minima and improve the meta-loss, we propose clipping the loss functions in the meta-gradient computation at the value found by the unperturbed parameters.
\begin{equation}
  \mathbb{E}_{\epsilon} \left[ \frac{\min \left[ L(\theta), L(\theta+\epsilon) \right] - \min \left[ L(\theta), L(\theta - \epsilon) \right]}{2\sigma^2} \right] \epsilon
\end{equation} 
where again $\epsilon \sim \mathcal{N}(0, I\sigma^2)$.

Intuitively, this biases against directions of high curvature in meta-optimization and empirically showed improved results. This strategy has the added benefit of heavily clipping the gradients of examples where loss spikes when atoms become too close, at the cost of an additional meta-loss calculation for $L(\theta)$.\footnote{In practice, this does not require a 50\% increase in meta-training time due to parallelization. On V100 GPUs, the increase in training time was as small as $15\%$. }

\paragraph{Genetic Algorithms (GA)} \mbox{}\\
Instead of relying on approximate meta-gradients, a simpler strategy perhaps is to adopt the perturbed parameters when they improve the meta-loss on a batch of random examples \cite{holland1992genetic,goldberg1988genetic}. To match the number of estimators of the meta-gradient used in ESMC, we use a population of size 80. At the end of each outer loop, the best performing parameters $\theta$ are kept and used for creating the next population by drawing from $\mathcal{N}(\theta, I\sigma^2)$ where $\sigma = 0.1$. By default, $\theta$ is kept constant when all samples perform worse than the baseline.

\begin{figure}[tb]
    \centering
    \includegraphics[width=0.45\textwidth]{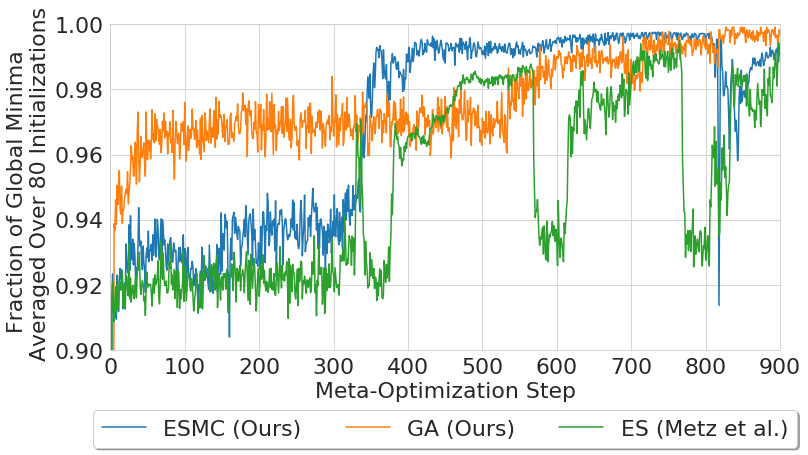}
    \vspace{-0.5em}
    \caption{Comparison of training strategies for learned optimizers on Lennard-Jones clusters highlights the need for ESMC or GA. Here, we see that our learned optimizers improve performance with respect to stability and averaged meta-loss.
    }
    \vspace{-0.5em}
    \label{fig:train_opt}
\end{figure}

\paragraph{Comparison of Methods} \mbox{}\\
A comparison of these strategies on a simplified learned optimizer setup is shown in Figure \ref{fig:train_opt}. The genetic-algorithm approach shows improvement early in meta-training which steadily converges to an optimizer where almost all initialization find the global minimum. In contrast, both ES and ESMC show a distinct transition in behavior around steps 300-400, which demarcates a transition from simple-to-learn descent behavior to more complex global minima discovery. The ESMC method is able to retain this behavior throughout meta-optimization, whereas traditional ES appears unstable and \textit{forgets}. Overall, both learned optimizer modifications show significant improvements in convergence speed and stability when compared to vanilla ES.
Details for this experiment can be found in Appendix \ref{app:comparison}. 

\label{sec:comparison}

\subsection{Additional Details}
In order to provide consistent scaling when averaging meta-losses across tasks, we divide energies by the best minimum found from applying Adam to 150 random initializations. This normalizes all losses to so that $-1$ is the best minimum found by Adam, ensuring that optimizers trained on multiple particle counts are not biased towards larger systems where the energy scales are lower. For each inner-loop, we apply 50000 optimization steps before computing meta-gradients. Batched training occurs with 80 random initializations of atomic structure problems. Once meta-gradients are averaged, a central controller meta-updates the parameters of the learned optimizer via Adam with a learning rate of $10^{-2}$ (which decays exponentially by 0.98 every 10 steps). This repeats for a total of 1000 meta-updates.

Finally, as the learned optimizer training does not guarantee a local minimum is found at the end of an optimization trajectory, we add 1000 steps of GD at learning rate of 0.001. We evaluate all strategies using 150 random initializations and report the mean and minimum energies found.

\subsection{Implementation Details}
The aforementioned potentials are coded using JAX-MD \cite{Schoenholz_Jax}. The learned optimizers are built in JAX \cite{Bradbury_Jax} to take advantage of automatic differentiation and vectorization of the optimization simulation. The associated training and evaluation utilized V100 GPUs. For distributed training, the controller batches computation on up to 8 GPUs.

\section{Experimental Results - Single Atom}
\begin{figure}[t]
    \centering
    \includegraphics[width=0.45\textwidth]{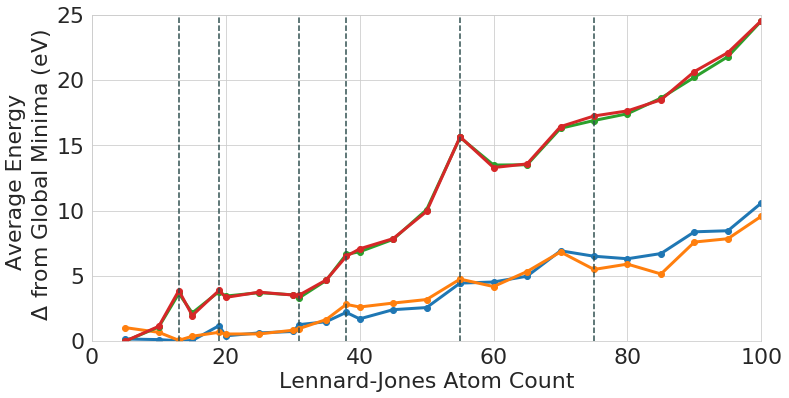}
    \includegraphics[width=0.45\textwidth]{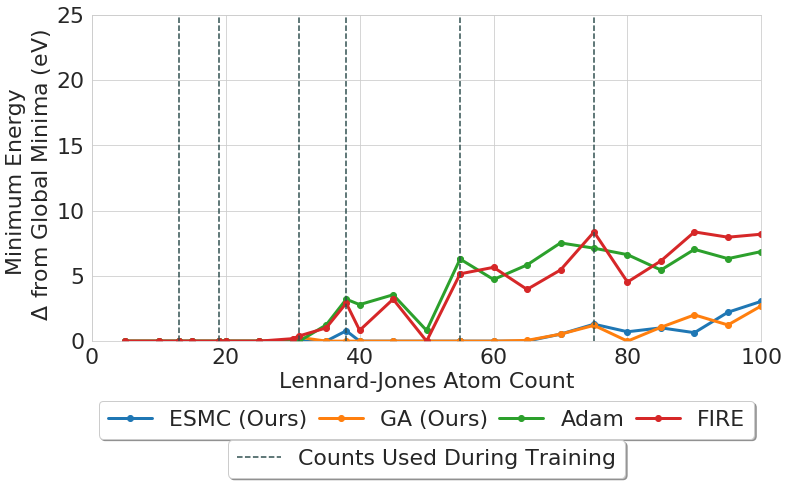}
    \vspace{-0.5em}
    \caption{Comparison of the learned optimizers and baseline methods on Lennard-Jones clusters. The learned optimizers are trained on a subset of the atom counts (demarcated by dashed lines) and evaluated via 150 random initializations. When compared to Adam and FIRE, the learned optimizers generically improve both the average energy per initialization (top) and best minima found (bottom) on atom counts unseen during training.}
    \vspace{-2.0em}
    \label{fig:lj_results}
\end{figure}

We start with a simplified set of potentials comprised of a single type of atom. For Lennard-Jones, we present results when a model is trained on a diverse set of atom counts, specifically \{13, 19, 31, 38, 55, 75\}, which helps stabilize learned optimizer training and improve generalization beyond the training set. In the results, starting with Figure \ref{fig:lj_results}, the learned optimizer shows significant performance gains when compared to the benchmark optimization algorithms of Adam and FIRE. This improvement not only takes the form of better minima but also better average energy per initialization. Note, the dotted lines correspond to the atom counts used during training; that the learned optimizers perform better between these lines shows that these learned optimizers generalize to tasks unseen during training yielding significant improvements over Adam and FIRE. Furthermore, the models generalize beyond the training distribution to tasks of up to 100 atoms.

\begin{figure}[t]
    \centering
    \includegraphics[width=0.42\textwidth]{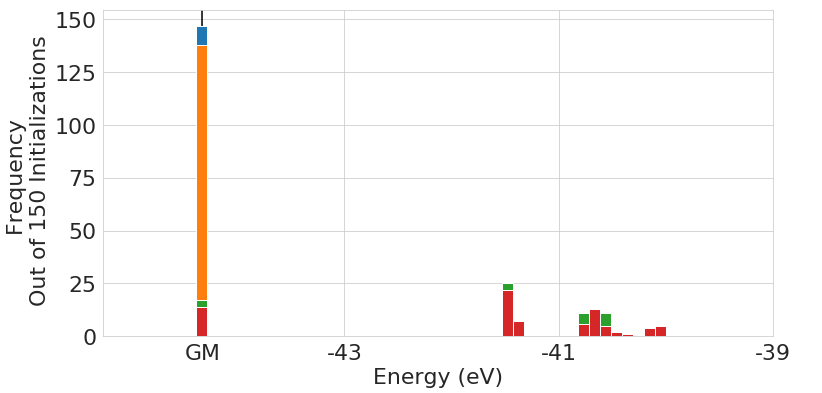}
    \includegraphics[width=0.42\textwidth]{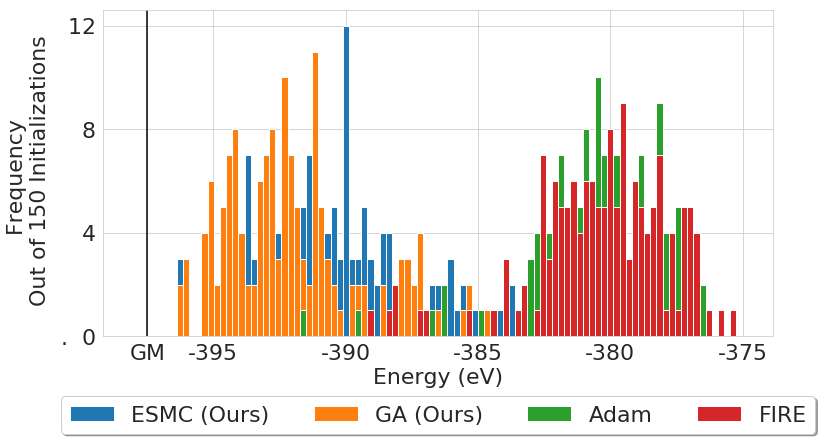}
    \vspace{0.5em}
    \caption{Distribution of minima found when baseline optimizers and the learned models are applied to 150 random initializations. For the Lennard-Jones task with 13 atoms (top), the learned optimizer find the global minimum in approximately 140 out of 150 trials, significantly better than the 20 of Adam and FIRE. Similar improvements are seen for the Lennard-Jones task with 75 atoms (bottom) where learned optimizers improve the minima found.}
    \vspace{-1.5em}
    \label{fig:lj_distribution}
\end{figure}

Figure \ref{fig:lj_distribution} analyzes the distribution of minima in greater detail for two canonical tasks: the 13 and 75 atom Lennard-Jones systems. The loss surface of the former is best described as a ``funnel" and even traditional algorithms can find the global minimum in about 20/150 random initializations. On the other hand, the 75 atom Lennard-Jones system has a glassy, optimization landscape, where high energy barriers exist between local minima \cite{Wales_Global} and the global minimum is difficult to find.

Interestingly, we first find that the baselines of Adam and FIRE yield similar performance per task after extensive hyper-parameter tuning.\footnote{While this trend was common in our experiments, additional work would be necessary to thoroughly compare these optimizers.} Nevertheless, both of our learned optimizer models show significant progress. With 13 atoms, the learned optimizers drastically increase the rate of global minimum discovery from 20/150 to above 140/150. With 75 atoms, the learned optimizers shift the distribution of minima found, finding minima within 1 eV of the global minimum compared to the 7 eV for Adam and FIRE. 

Similar results were obtained when learned optimizers were extended to the Gupta or SW potentials, when modeling 55-atom gold clusters and 64-atom silicon crystals respectively. Table \ref{tab:all_results} shows that the learned optimizers routinely surpass Adam and FIRE baselines and outperform Basin Hopping in a step-matched comparison.\footnote{Step-matched refers to an equivalent number of inner optimization steps. For details, please see the Appendix \ref{app:basin}} For silicon crystals, we note that the large gap between the global minimum and energies found arises from the difficulty of optimizing 64 atoms; due to the size, the problem reduces to finding finding stable amorphous structures (low energy local minima) rather than the true global minimum \cite{barkema1996event}. 

\label{sec:single}

\begin{savenotes}
\begin{table*}[tb]
\caption{Learned optimizers show improvement across all tested potential types. For each model and evaluation, the average and minimum are computed across 150 random initializations. All energies are reported in units of eV, and GM denotes the global minimum energy.}
\label{tab:all_results}
\begin{center}
\begin{small}
\begin{sc}
\begin{tabular}{lcccccccc}
\toprule
& & & &  \multicolumn{3}{c}{Step-Matched Baselines} & \multicolumn{2}{c}{Learned Optimizer} \\
Potential & \shortstack{Evaluation \\ \# Atoms} & GM & Metric & Adam & FIRE & \shortstack[c]{Basin \\ Hopping\footnote{In this comparison, Basin Hopping is limited to 50000 steps.}} & ESMC & GA\\
\midrule
Lennard-Jones & 13 & -44.33 & Min  & \textbf{-44.33} & \textbf{-44.33} & \textbf{-44.33} & \textbf{-44.33 }& \textbf{-44.33 }\\
\rowcolor{Silver} \cellcolor{White} & \cellcolor{White}  & \cellcolor{White}  & Mean &  -40.58 & -40.45 & -43.49 & -44.26 & \textbf{-44.31} \\ 
& 75 & -397.49 & Min & -390.34 & -389.12 & -392.16 & \textbf{-396.24} & \textbf{-396.28} \\ 
\rowcolor{Silver} \cellcolor{White} & \cellcolor{White}  & \cellcolor{White} & Mean & -380.52 & -380.23 & -381.49 & -390.33 & \textbf{-390.92}   \\
\midrule
Gupta Gold & 55 & -181.89 & Min & -180.94 & -181.75 & \textbf{-181.89} & \textbf{-181.89} & \textbf{-181.89}  \\ 
\rowcolor{Silver} \cellcolor{White} & \cellcolor{White}  & \cellcolor{White} & Mean & -179.94 & -180.94 & -181.38 & -181.51 & \textbf{-181.61}   \\
\midrule
SW silicon & 64 & -277.22 & Min & -60.08 & -261.44 & -261.64 & -262.95 & \textbf{-264.17} \\
\rowcolor{Silver} \cellcolor{White} & \cellcolor{White}  & \cellcolor{White}  & Mean & -256.83 & -257.01 & -259.14 & -260.14 & \textbf{-261.81} \\
\bottomrule
\end{tabular}
\end{sc}
\end{small}
\end{center}
\vspace{-2.0em}
\end{table*}
\end{savenotes}

\section{Behavioral Analysis}
\label{sec:analysis}
As the update function is parameterized by a neural network, it is unclear how learned optimizers improve atomic structural prediction. To explore the behavior, we provide three analyses showcasing an emergent `hopping' behavior and what features are critical for learned optimizer performance.

\subsection{Loss Trajectories}
\begin{figure}[tb]
    \hspace{\textwidth}
    \centering
    \includegraphics[width=0.40\textwidth]{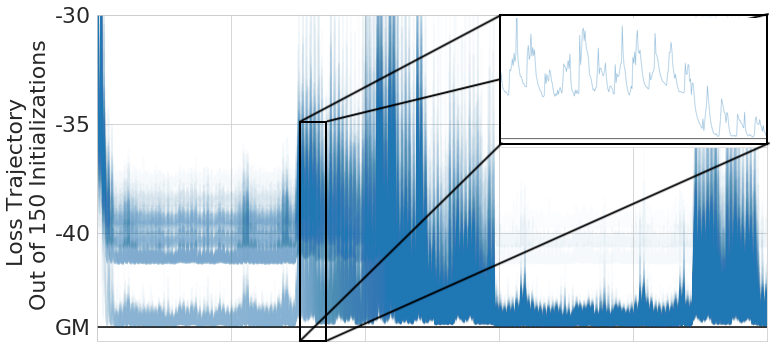}
    \includegraphics[width=0.40\textwidth]{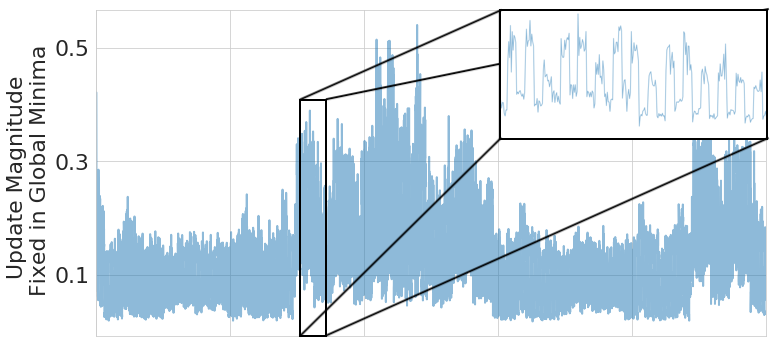}
    \includegraphics[width=0.40\textwidth]{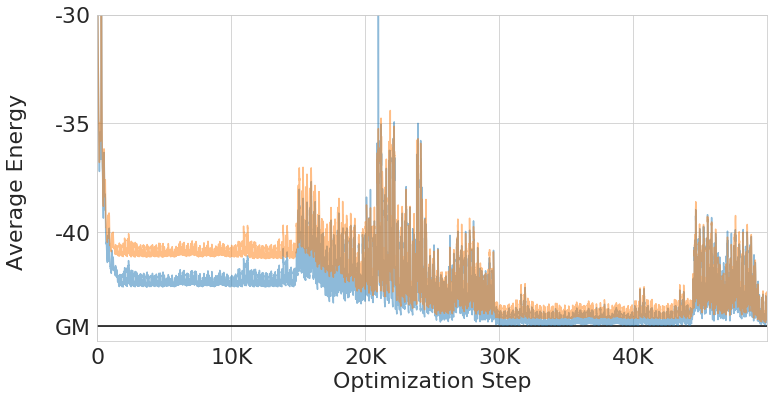}
    \includegraphics[width=0.40\textwidth]{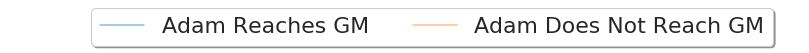}
    \vspace{-0.25em}
    \caption{Behavioral analyses of learned optimizers find that examples `hop' between basins rather than descent behavior. This is seen in the behavior of individual trajectories (top), where each trajectory has an opacity of 0.02 (so darker regions corresponds a greater number of examples). These `hops' are further supported by the spiking update magnitudes when fixed to a local minimum, suggesting that learned optimizers prioritize exploration of various basins (middle). For both of these diagrams, we zoom-in on a single trajectory, showing how the `hops' arise from large updates, followed by periods of descent. We also find that this hopping behavior corrects unlucky, random initializations that would not find the global minimum via Adam (bottom).}
    \vspace{-0.25em}
    \label{fig:trajectories}
\end{figure}

In Figure \ref{fig:trajectories} (top), we show loss trajectories when the learned optimizer is applied to the Lennard-Jones task with 13 atoms. Interestingly, the behavior of the loss function is not monotonic. While the model does rapidly descend into individual basins, many of these models display spikes in loss or `hopping' behavior where the model transitions between basins of different local minima at an erratic interval. More over, the optimizers have discovered characteristics that determine when to leave their basin. Figure \ref{fig:trajectories} (bottom) explores these trajectories in greater details, by filtering the `lucky' intializations that lead to the correct global minimum via Adam only. In cases where the parameters start in the correct basin, the learned optimizers performs better, acting like traditional Adam. For worse random starts, the learned optimizer will descend and then `hop' between basins.

\subsection{Behavior at Minima}

This `hopping' behavior appears to be key to the learned optimizer performance. Inspired by \citet{maheswaranathan2021reverse}, we fix the positions at the global minimum and compute the learned optimizer updates over an entire optimization trajectory. This strategy removes the influence of gradients in the learned optimizer (as they are zero) and helps visualize behavior as a function of step number. In Figure \ref{fig:trajectories} (middle), repeatedly applying the learned optimizer and measuring the update magnitude displays these `hops,' indicating that the model emphasizes on exploration and hopping between basins midway through these optimization trajectories (despite not receiving gradient signal to do so).

\subsection{Feature Importance}
Table \ref{tab:ablation} provides an ablation study to explain what input features to the learned optimizer are most important. To clarify the difference in performance, results are presented for a learned optimizer trained only on the 75 atom Lennard-Jones system (similar results were found for other clusters and crystals). To account for training instability, each result is the median over 10 shortened runs of 650 meta-updates.

We first see slight improvements coming from the addition of exponential moving averages of the gradient, similar to the benefits of momentum in stochastic optimization. However, what is most critical to model performance is the training step sine features and the radial symmetry functions. The sine features encode the optimization step via sine waves of various timescales (\{1, 3, 10, 30, 100, 300, 1000, 3000, 10000\}), and 
we hypothesize that these are helpful as they provides signal for when the learned optimizer should perform exploration and exploitation. Otherwise, the models tend to learn monotonic behavior that is similar in spirit to the Adam solutions. Finally, the newly introduced radial symmetry features also provide significant improvement, suggesting that per-position optimization is sub-optimal and rather information about particle neighbors (other than just gradients) is informative for optimization.

\begin{table}[tb]
\caption{Improvement arising from additional features shows that \begin{sc} Sine \end{sc} and \begin{sc} Radial \end{sc} features boost  model performance, suggesting 
information about step count and local neighborhoods of atoms are helpful for optimization. Results are the median performance out of 10 random training seeds. Note results are worse than Table \ref{tab:all_results} due to shortened training schedules. Lower is better.
}
\label{tab:ablation}
\vskip 0.15in
\begin{center}
\begin{small}
\begin{sc}
\setlength\tabcolsep{1.5pt}
\begin{tabular}{lcc}
\toprule
Optimizers & \multicolumn{2}{c}{Minimum Energy (eV)} \\
\midrule
\textbf{Baseline} \\
Adam & -390.3 & -390.3\\
\midrule
& $\Delta$ \text{ from Adam} & $\Delta$ \text{ from Adam} \\ 
\textbf{Learned}\;\textbf{Optimizer} & ESMC (ours) & GA (ours) \\
(1) Gradients & +0.2 & -0.8 \\
(2) Positions & +0.2 & -1.2 \\
(3) Decays & -2.0 & -2.0 \\
(4) Adam-Like & -1.0 & -2.8 \\
(5) Singular & -1.0 & -2.9 \\
(6) Species & -0.5 & -2.4 \\
(7) Sine & -2.3 & -3.6 \\
\midrule
(8) Radial & -3.4 & -4.4 \\
\bottomrule
\end{tabular}
\end{sc}
\end{small}
\end{center}
\vspace{-1.5em}
\end{table}

Overall, the behavioral analyses show that `hopping' behavior is critical to the performance of learned optimizers; however, given that an equivalent number of traditional Basin Hopping steps yields worse performance suggests more complex behavior also occurs.

\section{Experimental Results - Bimetallic Clusters}
\label{sec:bimetallic}
Having found that learned optimizers perform well in the case of single atom systems, we introduce additional complexity and explore generalization performance of the learned optimizers using bimetallic clusters. These systems are particularly interesting as purely gradient-based optimization methods such as Adam or FIRE fail, unable to pass the large energy barriers between local minima.
These potentials also allow for exploration of whether the learned behavior can transfer, a promising sign for the usage of these models in material design or crystal discovery.

\begin{figure}[tb]
    \centering
    \includegraphics[width=0.44\textwidth]{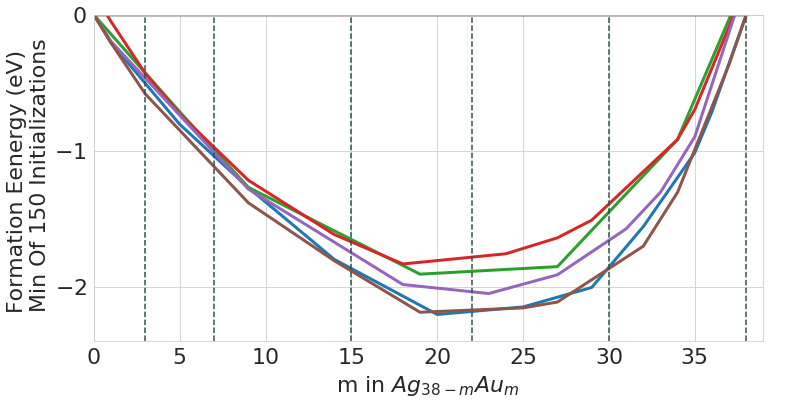}
    \includegraphics[width=0.44\textwidth]{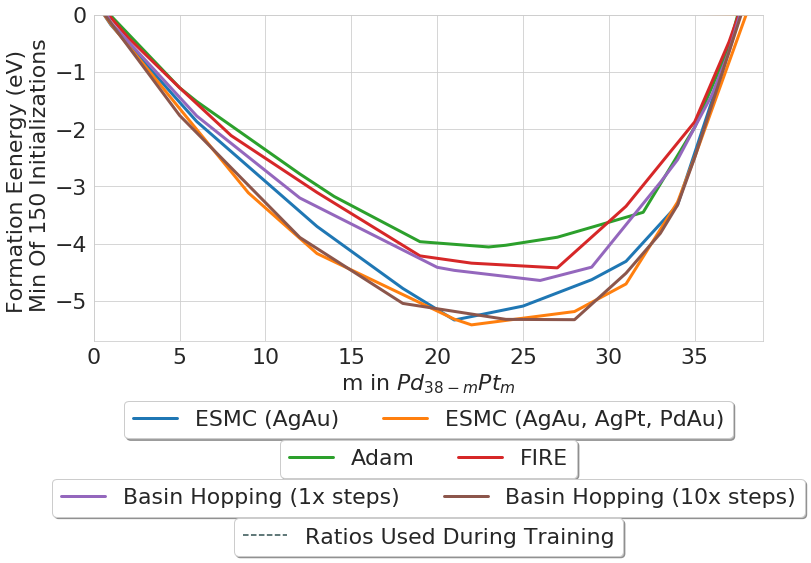}
    \vspace{-0.25em}
    \caption{Results for the bimetallic AgAu (top) and PdPt (bottom) clusters. For ESMC (AgAu), the learned optimizer is trained on only a subset of the possible ratios between Ag and Au. ESMC (AgAu, AgPt, PdAu) is trained only with AgAu, AgPt, PdAu clusters. On both AgAu and PdPt systems, both of our learned optimizers significantly outperform the baselines of Adam, FIRE, and step-matched Basin Hopping, which shows that the learned optimizers can generalize to new ratios or combinations of seen elements and new elements entirely unseen during training.}
    \vspace{-0.25em}
    \label{fig:agau}
\end{figure}

For the bimetallic clusters, we focus on the Gupta potential, whose parameters are modified to correspond to specific pairwise interactions \cite{Gupta}. The constant values used can be found in Appendix \ref{app:gupta}. First, the results of training learned optimizers on bimetallic clusters comprising of Silver (Ag) and Gold (Au) are shown in Figure \ref{fig:agau} (top). Fixing the total number of atoms at 38, we train the learned optimizer on $\text{Ag}_{38-m}\text{Au}_{m}$ for $m \in \{3, 7, 15, 22, 30, 38\} $ and test on all values of $m$. We present the convex hull of the formation energies of the clusters, as in equilibrium when excess silver and gold particles are present, only these clusters will be stable. The graphs show the robust empirical performance of learned optimizers, significantly outperforming Adam, FIRE, and a step-matched Basin Hopping benchmark. Only after 10x evaluation steps does Basin Hopping compete with the performance of the learned optimizer. This result indicates that the learned optimizers generalize as few of the AgAu clusters were used in training.

In the context of material design and crystal discovery, another core question is whether the learned optimizers will generalize beyond the set of atoms used for training. In Figure \ref{fig:agau} (bottom) we show the results from both the AgAu model described above and a second model trained on AgAu, AgPt, and PdAu. For both, we test on clusters of PdPt, which is not in the training set of either model. The learned optimizers show successful transfer performance, exceeding the step-matched Basin Hopping results. Only after 10x the number of evaluation steps can Basin Hopping compete with the learned optimizers (even after tuning, see Appendix \ref{app:basin}. While increasing the diversity of training tasks does improve generalization performance, both optimizers show an ability to transfer to unseen elements or combinations, a promising sign for this strategy of learning to optimize.\footnote{The improvement in step count only occurs after training the learned optimizer, which is expensive. The transference results, however, are promising, as this cost could be offset by the optimizer being used to efficiently evaluate a variety of element pairs.}

\section{Conclusion}
With current optimization techniques in material design and chemical physics requiring hand-crafted features or significant evaluation time, this paper explores the idea of global minimum discovery using learned optimizers. Although novel adaptations are required from the current state-of-the-art learned optimizers, we show that the resulting models can beat current baseline optimization techniques such as Adam and FIRE, not only in terms of minima discovery but also in terms of average energy per initialization. Better yet, these learned optimizers show signs of transference between potentials of similar varieties (ex: clusters parameterized by the Lennard-Jones and Gupta potentials), even on never before seen elements or combinations. Although a single optimizer for all task remains a goal for future work, learned optimizers show promise in automatically finding minima in complex optimization landscapes. We hope that the resulting models can aid in the design of new materials (e.g. for addressing energy challenges). 

\medskip

\section*{Acknowledgements}
We thank Jascha Sohl-dickstein and the Google Brain team for their help with the project.

\bibliography{example_paper}

\begin{thebibliography}{57}
\providecommand{\natexlab}[1]{#1}
\providecommand{\url}[1]{\texttt{#1}}
\expandafter\ifx\csname urlstyle\endcsname\relax
  \providecommand{\doi}[1]{doi: #1}\else
  \providecommand{\doi}{doi: \begingroup \urlstyle{rm}\Url}\fi

\bibitem[Abebe \& Solomatine(1998)Abebe and Solomatine]{Abebe_Application}
Abebe, A. and Solomatine, D.
\newblock Application of global optimization to the design of pipe networks.
\newblock In \emph{Proc. 3rd International Conference on Hydroinformatics,
  Copenhagen}, pp.\  989--996, 1998.

\bibitem[Andrychowicz et~al.(2016)Andrychowicz, Denil, Gomez, Hoffman, Pfau,
  Schaul, Shillingford, and De~Freitas]{Andrychowicz_Learning}
Andrychowicz, M., Denil, M., Gomez, S., Hoffman, M.~W., Pfau, D., Schaul, T.,
  Shillingford, B., and De~Freitas, N.
\newblock Learning to learn by gradient descent by gradient descent.
\newblock In \emph{Advances in neural information processing systems}, pp.\
  3981--3989, 2016.

\bibitem[Artrith et~al.(2013)Artrith, Hiller, and Behler]{Artrith_Neural}
Artrith, N., Hiller, B., and Behler, J.
\newblock Neural network potentials for metals and oxides--first applications
  to copper clusters at zinc oxide.
\newblock \emph{physica status solidi (b)}, 250\penalty0 (6):\penalty0
  1191--1203, 2013.

\bibitem[Barkema \& Mousseau(1996)Barkema and Mousseau]{barkema1996event}
Barkema, G. and Mousseau, N.
\newblock Event-based relaxation of continuous disordered systems.
\newblock \emph{Physical review letters}, 77\penalty0 (21):\penalty0 4358,
  1996.

\bibitem[Beeman(1976)]{Beeman_Some}
Beeman, D.
\newblock Some multistep methods for use in molecular dynamics calculations.
\newblock \emph{Journal of Computational Physics}, 20\penalty0 (2):\penalty0
  130 -- 139, 1976.
\newblock ISSN 0021-9991.
\newblock \doi{https://doi.org/10.1016/0021-9991(76)90059-0}.
\newblock URL
  \url{http://www.sciencedirect.com/science/article/pii/0021999176900590}.

\bibitem[Behler \& Parrinello(2007)Behler and Parrinello]{Behler_Generalized}
Behler, J. and Parrinello, M.
\newblock Generalized neural-network representation of high-dimensional
  potential-energy surfaces.
\newblock \emph{Physical review letters}, 98\penalty0 (14):\penalty0 146401,
  2007.

\bibitem[Bengio et~al.(1992)Bengio, Bengio, Cloutier, and
  Gecsei]{bengio1992optimization}
Bengio, S., Bengio, Y., Cloutier, J., and Gecsei, J.
\newblock On the optimization of a synaptic learning rule.
\newblock In \emph{Preprints Conf. Optimality in Artificial and Biological
  Neural Networks}, pp.\  6--8. Univ. of Texas, 1992.

\bibitem[Biswas \& Hamann(1986)Biswas and Hamann]{Biswas_Simulated}
Biswas, R. and Hamann, D.~R.
\newblock Simulated annealing of silicon atom clusters in langevin molecular
  dynamics.
\newblock \emph{Phys. Rev. B}, 34:\penalty0 895--901, Jul 1986.
\newblock \doi{10.1103/PhysRevB.34.895}.
\newblock URL \url{https://link.aps.org/doi/10.1103/PhysRevB.34.895}.

\bibitem[Bitzek et~al.(2006)Bitzek, Koskinen, G\"ahler, Moseler, and
  Gumbsch]{Bitzek_Structural}
Bitzek, E., Koskinen, P., G\"ahler, F., Moseler, M., and Gumbsch, P.
\newblock Structural relaxation made simple.
\newblock \emph{Phys. Rev. Lett.}, 97:\penalty0 170201, Oct 2006.
\newblock \doi{10.1103/PhysRevLett.97.170201}.
\newblock URL \url{https://link.aps.org/doi/10.1103/PhysRevLett.97.170201}.

\bibitem[Bradbury et~al.(2018)Bradbury, Frostig, Hawkins, Johnson, Leary,
  Maclaurin, and Wanderman-Milne]{Bradbury_Jax}
Bradbury, J., Frostig, R., Hawkins, P., Johnson, M.~J., Leary, C., Maclaurin,
  D., and Wanderman-Milne, S.
\newblock {JAX}: composable transformations of {P}ython+{N}um{P}y programs.
\newblock 2018.
\newblock URL \url{http://github.com/google/jax}.

\bibitem[Cheon et~al.(2020)Cheon, Yang, McCloskey, Reed, and
  Cubuk]{cheon2020crystal}
Cheon, G., Yang, L., McCloskey, K., Reed, E.~J., and Cubuk, E.~D.
\newblock Crystal structure search with random relaxations using graph
  networks.
\newblock \emph{arXiv preprint arXiv:2012.02920}, 2020.

\bibitem[Choromanska et~al.(2015)Choromanska, Henaff, Mathieu, Arous, and
  LeCun]{choromanska2015loss}
Choromanska, A., Henaff, M., Mathieu, M., Arous, G.~B., and LeCun, Y.
\newblock The loss surfaces of multilayer networks.
\newblock In \emph{Artificial intelligence and statistics}, pp.\  192--204,
  2015.

\bibitem[Cubuk et~al.(2015)Cubuk, Schoenholz, Rieser, Malone, Rottler, Durian,
  Kaxiras, and Liu]{cubuk2015identifying}
Cubuk, E.~D., Schoenholz, S.~S., Rieser, J.~M., Malone, B.~D., Rottler, J.,
  Durian, D.~J., Kaxiras, E., and Liu, A.~J.
\newblock Identifying structural flow defects in disordered solids using
  machine-learning methods.
\newblock \emph{Physical review letters}, 114\penalty0 (10):\penalty0 108001,
  2015.

\bibitem[Doye et~al.(1995)Doye, Wales, and Berry]{Doye_Effect}
Doye, J.~P., Wales, D.~J., and Berry, R.~S.
\newblock The effect of the range of the potential on the structures of
  clusters.
\newblock \emph{The Journal of chemical physics}, 103\penalty0 (10):\penalty0
  4234--4249, 1995.

\bibitem[Doye et~al.(1999)Doye, Miller, and Wales]{Doye_Evolution}
Doye, J. P.~K., Miller, M.~A., and Wales, D.~J.
\newblock Evolution of the potential energy surface with size for lennard-jones
  clusters.
\newblock \emph{The Journal of Chemical Physics}, 111\penalty0 (18):\penalty0
  8417--8428, 1999.
\newblock \doi{10.1063/1.480217}.
\newblock URL \url{https://doi.org/10.1063/1.480217}.

\bibitem[Farges et~al.(1985)Farges, {De Feraudy}, Raoult, and
  Torchet]{Farges_Cluster}
Farges, J., {De Feraudy}, M., Raoult, B., and Torchet, G.
\newblock Cluster models made of double icosahedron units.
\newblock \emph{Surface Science}, 156:\penalty0 370 -- 378, 1985.
\newblock ISSN 0039-6028.
\newblock \doi{https://doi.org/10.1016/0039-6028(85)90596-5}.
\newblock URL
  \url{http://www.sciencedirect.com/science/article/pii/0039602885905965}.

\bibitem[Flikkema \& Bromley(2004)Flikkema and Bromley]{Flikkema_Dedicated}
Flikkema, E. and Bromley, S.~T.
\newblock Dedicated global optimization search for ground state silica
  nanoclusters.
\newblock \emph{The Journal of Physical Chemistry B}, 108\penalty0
  (28):\penalty0 9638--9645, 2004.
\newblock \doi{10.1021/jp049783r}.
\newblock URL \url{https://doi.org/10.1021/jp049783r}.

\bibitem[Giannozzi et~al.(2009)Giannozzi, Baroni, Bonini, Calandra, Car,
  Cavazzoni, Ceresoli, Chiarotti, Cococcioni, Dabo, et~al.]{Giannozzi_Quantum}
Giannozzi, P., Baroni, S., Bonini, N., Calandra, M., Car, R., Cavazzoni, C.,
  Ceresoli, D., Chiarotti, G.~L., Cococcioni, M., Dabo, I., et~al.
\newblock Quantum espresso: a modular and open-source software project for
  quantum simulations of materials.
\newblock \emph{Journal of physics: Condensed matter}, 21\penalty0
  (39):\penalty0 395502, 2009.

\bibitem[Gilmer et~al.(2017)Gilmer, Schoenholz, Riley, Vinyals, and
  Dahl]{gilmer2017neural}
Gilmer, J., Schoenholz, S.~S., Riley, P.~F., Vinyals, O., and Dahl, G.~E.
\newblock Neural message passing for quantum chemistry.
\newblock In \emph{International Conference on Machine Learning}, pp.\
  1263--1272. PMLR, 2017.

\bibitem[Goldberg \& Holland(1988)Goldberg and Holland]{goldberg1988genetic}
Goldberg, D.~E. and Holland, J.~H.
\newblock Genetic algorithms and machine learning.
\newblock 1988.

\bibitem[Gu et~al.(2019)Gu, Greydanus, Metz, Maheswaranathan, and
  Sohl-Dickstein]{gu2019meta}
Gu, K., Greydanus, S., Metz, L., Maheswaranathan, N., and Sohl-Dickstein, J.
\newblock Meta-learning biologically plausible semi-supervised update rules.
\newblock \emph{bioRxiv}, 2019.

\bibitem[Gupta(1981)]{Gupta}
Gupta, R.~P.
\newblock Lattice relaxation at a metal surface.
\newblock \emph{Phys. Rev. B}, 23:\penalty0 6265--6270, Jun 1981.
\newblock \doi{10.1103/PhysRevB.23.6265}.
\newblock URL \url{https://link.aps.org/doi/10.1103/PhysRevB.23.6265}.

\bibitem[Heek et~al.(2020)Heek, Levskaya, Oliver, Ritter, Rondepierre, Steiner,
  and van {Z}ee]{flax2020github}
Heek, J., Levskaya, A., Oliver, A., Ritter, M., Rondepierre, B., Steiner, A.,
  and van {Z}ee, M.
\newblock {F}lax: A neural network library and ecosystem for {JAX}, 2020.
\newblock URL \url{http://github.com/google/flax}.

\bibitem[Hoare \& Pal(1971)Hoare and Pal]{Hoare_Physical}
Hoare, M. and Pal, P.
\newblock Physical cluster mechanics: Statics and energy surfaces for monatomic
  systems.
\newblock \emph{Advances in Physics}, 20\penalty0 (84):\penalty0 161--196,
  1971.

\bibitem[Hohenberg \& Kohn(1964)Hohenberg and Kohn]{hohenberg1964inhomogeneous}
Hohenberg, P. and Kohn, W.
\newblock Inhomogeneous electron gas.
\newblock \emph{Physical review}, 136\penalty0 (3B):\penalty0 B864, 1964.

\bibitem[Holland(1992)]{holland1992genetic}
Holland, J.~H.
\newblock Genetic algorithms.
\newblock \emph{Scientific american}, 267\penalty0 (1):\penalty0 66--73, 1992.

\bibitem[Jones \& Chapman(1924)Jones and Chapman]{Jones}
Jones, J.~E. and Chapman, S.
\newblock On the determination of molecular fields.\&\#x2014;i. from the
  variation of the viscosity of a gas with temperature.
\newblock \emph{Proceedings of the Royal Society of London. Series A,
  Containing Papers of a Mathematical and Physical Character}, 106\penalty0
  (738):\penalty0 441--462, 1924.
\newblock \doi{10.1098/rspa.1924.0081}.
\newblock URL
  \url{https://royalsocietypublishing.org/doi/abs/10.1098/rspa.1924.0081}.

\bibitem[Kingma \& Ba(2014)Kingma and Ba]{Kingma_Adam}
Kingma, D.~P. and Ba, J.
\newblock Adam: A method for stochastic optimization.
\newblock \emph{arXiv preprint arXiv:1412.6980}, 2014.

\bibitem[Kirkpatrick et~al.(1983)Kirkpatrick, Gelatt, and
  Vecchi]{Kirkpatrick_Optimization}
Kirkpatrick, S., Gelatt, C.~D., and Vecchi, M.~P.
\newblock Optimization by simulated annealing.
\newblock \emph{science}, 220\penalty0 (4598):\penalty0 671--680, 1983.

\bibitem[LeCun et~al.(2012)LeCun, Bottou, Orr, and
  M{\"u}ller]{lecun2012efficient}
LeCun, Y.~A., Bottou, L., Orr, G.~B., and M{\"u}ller, K.-R.
\newblock Efficient backprop.
\newblock In \emph{Neural networks: Tricks of the trade}, pp.\  9--48.
  Springer, 2012.

\bibitem[Liwo et~al.(1999)Liwo, Lee, Ripoll, Pillardy, and
  Scheraga]{Liwo_Protein}
Liwo, A., Lee, J., Ripoll, D.~R., Pillardy, J., and Scheraga, H.~A.
\newblock Protein structure prediction by global optimization of a potential
  energy function.
\newblock \emph{Proceedings of the National Academy of Sciences}, 96\penalty0
  (10):\penalty0 5482--5485, 1999.
\newblock ISSN 0027-8424.
\newblock \doi{10.1073/pnas.96.10.5482}.
\newblock URL \url{https://www.pnas.org/content/96/10/5482}.

\bibitem[Luo et~al.(2018)Luo, Wu, and Lee]{Lee_No}
Luo, J., Wu, C., and Lee, J.
\newblock No spurious local minima in a two hidden unit relu network.
\newblock January 2018.
\newblock 6th International Conference on Learning Representations, ICLR 2018 ;
  Conference date: 30-04-2018 Through 03-05-2018.

\bibitem[Lv et~al.(2017)Lv, Jiang, and Li]{lv2017learning}
Lv, K., Jiang, S., and Li, J.
\newblock Learning gradient descent: Better generalization and longer horizons.
\newblock \emph{arXiv preprint arXiv:1703.03633}, 2017.

\bibitem[Maheswaranathan et~al.(2021)Maheswaranathan, Sussillo, Metz, Sun, and
  Sohl-Dickstein]{maheswaranathan2021reverse}
Maheswaranathan, N., Sussillo, D., Metz, L., Sun, R., and Sohl-Dickstein, J.
\newblock Reverse engineering learned optimizers reveals known and novel
  mechanisms, 2021.
\newblock URL \url{https://openreview.net/forum?id=y_pDlU_FLS}.

\bibitem[Metz et~al.(2018)Metz, Maheswaranathan, Nixon, Freeman, and
  Sohl-Dickstein]{Metz_Learned}
Metz, L., Maheswaranathan, N., Nixon, J., Freeman, D., and Sohl-Dickstein, J.
\newblock Learned optimizers that outperform sgd on wall-clock and test loss.
\newblock In \emph{Proceedings of the 2nd Workshop on Meta-Learning.
  MetaLearn}, 2018.

\bibitem[Metz et~al.(2019{\natexlab{a}})Metz, Maheswaranathan, Nixon, Freeman,
  and Sohl-Dickstein]{Metz_Understanding}
Metz, L., Maheswaranathan, N., Nixon, J., Freeman, D., and Sohl-Dickstein, J.
\newblock Understanding and correcting pathologies in the training of learned
  optimizers.
\newblock In \emph{International Conference on Machine Learning}, pp.\
  4556--4565, 2019{\natexlab{a}}.

\bibitem[Metz et~al.(2019{\natexlab{b}})Metz, Maheswaranathan, Shlens,
  Sohl-Dickstein, and Cubuk]{Metz_Using}
Metz, L., Maheswaranathan, N., Shlens, J., Sohl-Dickstein, J., and Cubuk, E.~D.
\newblock Using learned optimizers to make models robust to input noise.
\newblock \emph{arXiv preprint arXiv:1906.03367}, 2019{\natexlab{b}}.

\bibitem[Metz et~al.(2020)Metz, Maheswaranathan, Freeman, Poole, and
  Sohl-Dickstein]{metz2020tasks}
Metz, L., Maheswaranathan, N., Freeman, C.~D., Poole, B., and Sohl-Dickstein,
  J.
\newblock Tasks, stability, architecture, and compute: Training more effective
  learned optimizers, and using them to train themselves.
\newblock \emph{arXiv preprint arXiv:2009.11243}, 2020.

\bibitem[Michaelian et~al.(1999)Michaelian, Rendón, and Garzón]{Au}
Michaelian, K., Rendón, N., and Garzón, I.
\newblock Structure and energetics of ni, ag, and au nanoclusters.
\newblock \emph{Physical Review B}, 60, 07 1999.
\newblock \doi{10.1103/PhysRevB.60.2000}.

\bibitem[Oganov et~al.(2019)Oganov, Pickard, Zhu, and
  Needs]{oganov2019structure}
Oganov, A.~R., Pickard, C.~J., Zhu, Q., and Needs, R.~J.
\newblock Structure prediction drives materials discovery.
\newblock \emph{Nature Reviews Materials}, 4\penalty0 (5):\penalty0 331--348,
  2019.

\bibitem[Pascanu et~al.(2013)Pascanu, Mikolov, and
  Bengio]{pascanu2013difficulty}
Pascanu, R., Mikolov, T., and Bengio, Y.
\newblock On the difficulty of training recurrent neural networks.
\newblock In \emph{International conference on machine learning}, pp.\
  1310--1318. PMLR, 2013.

\bibitem[Paz-Borb{\'o}n et~al.(2008)Paz-Borb{\'o}n, Johnston, Barcaro, and
  Fortunelli]{PazBorbn2008StructuralMM}
Paz-Borb{\'o}n, L.~O., Johnston, R., Barcaro, G., and Fortunelli, A.
\newblock Structural motifs, mixing, and segregation effects in 38-atom binary
  clusters.
\newblock \emph{The Journal of chemical physics}, 128 13:\penalty0 134517,
  2008.

\bibitem[Pickard \& Needs(2006)Pickard and Needs]{Pickard_High}
Pickard, C.~J. and Needs, R.~J.
\newblock High-pressure phases of silane.
\newblock \emph{Phys. Rev. Lett.}, 97:\penalty0 045504, Jul 2006.
\newblock \doi{10.1103/PhysRevLett.97.045504}.
\newblock URL \url{https://link.aps.org/doi/10.1103/PhysRevLett.97.045504}.

\bibitem[Pickard \& Needs(2011)Pickard and Needs]{Pickard_Ab}
Pickard, C.~J. and Needs, R.~J.
\newblock Ab initiorandom structure searching.
\newblock \emph{Journal of Physics: Condensed Matter}, 23\penalty0
  (5):\penalty0 053201, jan 2011.
\newblock \doi{10.1088/0953-8984/23/5/053201}.
\newblock URL \url{https://doi.org/10.1088%2F0953-8984%2F23%2F5%2F053201}.

\bibitem[Rosato et~al.(1989)Rosato, Guillope, and Legrand]{Rosato}
Rosato, V., Guillope, M., and Legrand, B.
\newblock Thermodynamical and structural properties of f.c.c. transition metals
  using a simple tight-binding model.
\newblock \emph{Philosophical Magazine A}, 59\penalty0 (2):\penalty0 321--336,
  1989.
\newblock \doi{10.1080/01418618908205062}.
\newblock URL
  \url{https://www.tandfonline.com/doi/abs/10.1080/01418618908205062}.

\bibitem[Ryoo \& Sahinidis(1995)Ryoo and Sahinidis]{Ryoo_Global}
Ryoo, H.~S. and Sahinidis, N.~V.
\newblock Global optimization of nonconvex nlps and minlps with applications in
  process design.
\newblock \emph{Computers \& Chemical Engineering}, 19\penalty0 (5):\penalty0
  551--566, 1995.

\bibitem[Salimans et~al.(2017)Salimans, Ho, Chen, Sidor, and
  Sutskever]{salimans2017evolution}
Salimans, T., Ho, J., Chen, X., Sidor, S., and Sutskever, I.
\newblock Evolution strategies as a scalable alternative to reinforcement
  learning.
\newblock \emph{arXiv preprint arXiv:1703.03864}, 2017.

\bibitem[Schoenholz \& Cubuk(2019)Schoenholz and Cubuk]{Schoenholz_Jax}
Schoenholz, S.~S. and Cubuk, E.~D.
\newblock Jax, md: End-to-end differentiable, hardware accelerated, molecular
  dynamics in pure python.
\newblock \emph{arXiv preprint arXiv:1912.04232}, 2019.

\bibitem[Sch{\"u}tt et~al.(2017)Sch{\"u}tt, Kindermans, Sauceda, Chmiela,
  Tkatchenko, and M{\"u}ller]{schutt2017schnet}
Sch{\"u}tt, K., Kindermans, P.-J., Sauceda, H., Chmiela, S., Tkatchenko, A.,
  and M{\"u}ller, K.-R.
\newblock Schnet: a continuous-filter convolutional neural network for modeling
  quantum interactions.
\newblock In \emph{Proceedings of the 31st International Conference on Neural
  Information Processing Systems}, pp.\  992--1002, 2017.

\bibitem[Stillinger \& Weber(1985)Stillinger and Weber]{stillinger1985computer}
Stillinger, F.~H. and Weber, T.~A.
\newblock Computer simulation of local order in condensed phases of silicon.
\newblock \emph{Physical review B}, 31\penalty0 (8):\penalty0 5262, 1985.

\bibitem[Sutton(1993)]{sutton1993electronic}
Sutton, A.~P.
\newblock \emph{Electronic structure of materials}.
\newblock Clarendon Press, 1993.

\bibitem[Tran \& Johnston(2011)Tran and Johnston]{Tran_Study}
Tran, D. and Johnston, R.
\newblock Study of 40-atom pt-au clusters using a combined empirical
  potential-density functional approach.
\newblock \emph{Proceedings of The Royal Society A: Mathematical, Physical and
  Engineering Sciences}, 467:\penalty0 2004--2019, 05 2011.
\newblock \doi{10.1098/rspa.2010.0562}.

\bibitem[Tsai \& Jordan(1993)Tsai and Jordan]{Tsai_Use}
Tsai, C.~J. and Jordan, K.~D.
\newblock Use of an eigenmode method to locate the stationary points on the
  potential energy surfaces of selected argon and water clusters.
\newblock \emph{The Journal of Physical Chemistry}, 97\penalty0 (43):\penalty0
  11227--11237, 1993.
\newblock \doi{10.1021/j100145a019}.
\newblock URL \url{https://doi.org/10.1021/j100145a019}.

\bibitem[Wales \& Doye(1997)Wales and Doye]{Wales_Global}
Wales, D.~J. and Doye, J.~P.
\newblock Global optimization by basin-hopping and the lowest energy structures
  of lennard-jones clusters containing up to 110 atoms.
\newblock \emph{The Journal of Physical Chemistry A}, 101\penalty0
  (28):\penalty0 5111--5116, 1997.

\bibitem[Wichrowska et~al.(2017)Wichrowska, Maheswaranathan, Hoffman,
  Colmenarejo, Denil, de~Freitas, and Sohl-Dickstein]{Wichrowska_Learned}
Wichrowska, O., Maheswaranathan, N., Hoffman, M.~W., Colmenarejo, S.~G., Denil,
  M., de~Freitas, N., and Sohl-Dickstein, J.
\newblock Learned optimizers that scale and generalize.
\newblock \emph{arXiv preprint arXiv:1703.04813}, 2017.

\bibitem[Williams(1992)]{Williams_Simple}
Williams, R.~J.
\newblock Simple statistical gradient-following algorithms for connectionist
  reinforcement learning.
\newblock \emph{Mach. Learn.}, 8\penalty0 (3–4):\penalty0 229–256, May
  1992.
\newblock ISSN 0885-6125.
\newblock \doi{10.1007/BF00992696}.
\newblock URL \url{https://doi.org/10.1007/BF00992696}.

\bibitem[Zilka et~al.(2017)Zilka, Dudenko, Hughes, Williams, Sturniolo, Franks,
  Pickard, Yates, Harris, and Brown]{Zilka_Ab}
Zilka, M., Dudenko, D.~V., Hughes, C.~E., Williams, P.~A., Sturniolo, S.,
  Franks, W.~T., Pickard, C.~J., Yates, J.~R., Harris, K. D.~M., and Brown,
  S.~P.
\newblock Ab initio random structure searching of organic molecular solids:
  assessment and validation against experimental data.
\newblock \emph{Phys. Chem. Chem. Phys.}, 19:\penalty0 25949--25960, 2017.
\newblock \doi{10.1039/C7CP04186A}.
\newblock URL \url{http://dx.doi.org/10.1039/C7CP04186A}.

\end{thebibliography}
\bibliographystyle{icml2021}

\newpage

\clearpage
\appendix

\section{Empirical Potentials}
Empirical potentials are approximations of the potential energy surface for physical systems. In contrast to Density-Functional Theory which computes energies from first principles \cite{hohenberg1964inhomogeneous}, empirical potentials are generic functional forms whose parameters are fitted to correspond to individual systems and which are designed to be efficient to compute. In our paper, these empirical potentials enable the training of learned optimizers, as we can easily batch computation and use automatic differentiation techniques to quickly calculate gradients/forces.

Moreover, the minima of  empirical potentials are likely to generalize to those found by more-accurate computations, suggesting that the learned optimizers trained on these approximations will generalize as well. In this paper, we study a few functional forms of empirical potentials:

\subsection{Lennard-Jones Clusters}
First, the Lennard-Jones Clusters are the archetype of a simple-to-compute potential and are often used to model spherically-symmetric particles or atoms in free-space (a perfect vacuum with no other particles in the entire system). For example, this empirical potential can be used to model nobel gasses or methane \cite{Jones}. The energy landscape is defined only by pairwise distances between particles, denoted as $d_{ij}$ for atoms $i,j$.

\begin{equation}
    \sum_i \sum_{j>i} \epsilon \left[ \left( \frac{d_{0}}{d_{ij}} \right)^{12} - 2 \left( \frac{d_{0}}{d_{ij}} \right)^6 \right]
\end{equation}

where $\epsilon$ describes the minimum two-particle energy and $d_{0}$ describes the distance where this occurs. Following prior work, in our paper we set $\epsilon$ and $d_{0}$ to 1, as the resulting minima structures can be scaled as necessary for systems where these settings do not hold.

While this model of atoms may seem simple, the corresponding optimization problem is anything but. For the corresponding task with 75 atoms, common gradient-based techniques such as Adam or FIRE cannot obtain the global minimum value and even the best trial out of 150 random initializations has an error greater than 7eV. Please see the main body of the paper for additional discussion of the difficulties of optimizing Lennard-Jones Clusters.  

\subsection{Gupta Clusters}
\label{app:gupta}

The Gupta empirical potential adds an additional layer of complexity when modelling the energy of atomic structures. Designed to model lattice relaxations at a metal surface, the Gupta model provides an improved approximation by including a second-moment estimate of the tight-binding Hamiltonian \cite{Gupta}. The Gupta potential has been widely used in studying and predicting the stable structures of noble metals and bimetallic clusters.

The functional form of the Gupta potential is as follows:
\begin{align}
    \sum_{i} &\sum_{j > i} A \exp{\left[p \left(1 - \frac{d_{ij}}{d_0}\right) \right]} \nonumber \\
    &- \mathcal{\xi} \sum_{i} \sqrt{\sum_{j>i} \exp{\left[2q\left(1 - \frac{d_{ij}}{d_0}\right)\right]}}
\end{align}

The values $d_0, A, \xi, p, q$ are parameters of the Gupta potential that describe specific  inter-particle interactions. In contrast to the Lennard-Jones system, these values cannot be factored out or set to defaults. Instead, the parameters are dervied either experimentally or by fitting to Density-Functional Theory data from the bulk-faced cubic systems. Note, when the system consists of only one type of atom, the parameters are the same for all pairs of particles. However, for bimetallic clusters with multiple types of atoms, these parameters can depend on the type of interaction. For example, in Ag-Au clusters, there will be 3 possible values for each of these constants corresponding to Ag-Ag, Ag-Au, and Au-Au interactions.

As the minima structures of the Gupta potential can vary based on the exact parameter values used, all of our experiments are based on the already-discovered configurations by \citet{PazBorbn2008StructuralMM}. Table \ref{tab:gupta_parameters} provides the parameters used in our single-atom experiment modelling Au (gold) with 55 atoms and bimetallic experiments of Ag-Au, Ag-Pt, Pd-Au, and Pd-Pt clusters.

\subsection{Stillinger-Weber (SW)}
\label{app:sw}
Stillinger-Weber potentials are designed to provide more accurate estimations of semiconductors and do so by including a three-body angular term (i.e. penalizing for deviations from an optimal angle within the crystal structure). In our paper, we use the Stillinger-Weber potential to model Silicon crystals. This system is distinct from the other two benchmarks, as the crystal model assumes the lattice structure is repeatedly infinitely in space (although a cutoff in interaction distance allows for models to use finite tilings).

For this system, the energy is defined by:
\newpage

\begin{center}
\begin{equation}
    \sum_{i} \sum_{j > i} \theta_2(d_{ij}) + \sum_{i}\sum_{j \neq i} \sum_{k > j} \theta_3(d_{ij}, d_{ik}, \theta_{ijk}) 
\end{equation}
\begin{align*}
    \theta_2(d_{ij}) = A\epsilon \left[B \left( \frac{\sigma}{d_{ij}} \right)^p - \left( \frac{\sigma}{d_{ij}} \right)^q \right] \exp{\left[ \frac{\sigma}{d_{ij} - a \sigma} \right]} \\
    \theta_3(d_{ij}, d_{ik}, \theta_{ijk}) = \lambda \epsilon \left( \cos {\theta_{ijk}} - \cos{\theta_{0}} \right)^2 \exp{\left[ \frac{\gamma \sigma}{d_{ij} - a\sigma} \right]}
\end{align*}
\end{center} 
where again the scalar parameters are fit to the system being studied. For modelling Silicon, the parameters used in our experiments are provided in Table \ref{tab:sw_parameters}.

Additionally, we only study the optimization problem of a simple cubic lattice structure, where the lattice vectors are defined by $\Vec{i}, \Vec{j}, \Vec{k}$. Cells of 8 atoms are initialized to have size 5.248 \AA, and there is no lattice vector optimization during the course of training. Furthermore, we define the cutoff for atomic interactions to be 3.77 \AA.

\begin{table}[h!]
\caption{Gupta Potential Coefficients}
\label{tab:gupta_parameters}
\vskip 0.15in
\begin{center}
\begin{small}
\begin{sc}
\begin{tabular}{llllll}
\toprule
 & p & q & $d_{0}$ & A & $\xi$ \\
\midrule
\multicolumn{6}{l}{\textbf{Au (Gold) 55}} \\
Au-Au & 10.229 & 4.036 & 2.884 & 1.790 & 0.2061 \\
\hline
\multicolumn{6}{l}{\textbf{Bimetallic Ag-Au}} \\
Ag-Ag & 10.85 & 3.18 & 2.8921 & 1.1895 & 0.1031 \\
Ag-Au & 10.494 & 3.607 & 2.8885 & 1.4874 & 0.1488 \\
Au-Au & 10.139 & 4.033 & 2.885 & 1.8153 & 0.2096 \\
\hline
\multicolumn{6}{l}{\textbf{Bimetallic Ag-Pt}} \\
Ag-Ag & 10.86 & 3.18 & 2.8921 & 1.1895 & 0.1031\\
Ag-Pt & 10.73 & 3.57 & 2.833 & 1.79 & 0.175 \\
Pt-Pt & 10.612 & 4.004 & 2.7747 & 2.695 & 0.2975 \\
\hline
\multicolumn{6}{l}{\textbf{Bimetallic Pd-Au}} \\
Pd-Pd & 10.867 & 3.742 & 2.7485 & 1.718 & 0.1746 \\
Pd-Au & 10.54 & 3.89 & 2.816 & 1.75 & 0.19\\
Au-Au & 10.299 & 4.036 & 2.884 & 1.79 & 0.2061 \\
\hline
\multicolumn{6}{l}{\textbf{Bimetallic Pd-Pt}} \\
Pd-Pd & 10.867 & 3.742 & 2.7485 & 1.718 & 0.1746\\
Pd-Pt & 10.74 & 3.87 & 2.76 & 2.2 & 0.23\\
Pt-Pt & 10.612 & 4.004 & 2.7747 & 2.695 & 0.2975 \\

\bottomrule
\end{tabular}
\end{sc}
\end{small}
\end{center}
\vskip -0.1in
\end{table}

\begin{table}[t]
\caption{Stillinger-Weber Potential Coefficients}
\label{tab:sw_parameters}
\vskip 0.15in
\begin{center}
\begin{small}
\begin{sc}
\addtolength{\tabcolsep}{-2pt}
\begin{tabular}{lllllllll}
\toprule
& A & $\epsilon$ & B & p & $\lambda$ & $\gamma$ & $\sigma$ & a \\
\midrule
\multicolumn{6}{l}{\textbf{Silicon Crystals}} \\
Si-Si & 7.049 &  2.168 & 0.602 & 4 & 21.0 & 1.2 & 2.0951 & 1.8 \\
\bottomrule
\end{tabular}
\addtolength{\tabcolsep}{6pt}
\end{sc}
\end{small}
\end{center}
\vskip -0.1in
\end{table}

\section{Comparison of Training Strategies}
\label{app:comparison}
In order to compare the meta-training strategies of ES, ESMC (ours) and GA (ours), we focus on a simplified set-up of the learned optimizers. Specifically, to remove the noise originating from the meta-training on diverse tasks, we focus on models trained only on the 13-atom Lennard-Jones clusters. As the model is trained only on 1 task, we only need to train for 900 meta-updates before both ESMC (ours) and GA (ours) appear to reach the global minimum on almost every single initialization. In contrast, traditional ES appears to be unstable, deviating greatly in the meta-training loss.

As the loss landscape of this system is `funneled' and a low-energy paths exist between the local minima and the global minimum, we do not expect sporadic behavior of the learned optimizer to arise from the optimization problem itself. Instead, the erratic behavior of appears to be coming from instability in training, which is solved by our ESMC and GA training strategies.

\section{Baseline Optimizer Tuning}
\label{app:adam}
\label{app:fire}
The baseline optimization techniques used in this paper (Adam, FIRE, Basin Hopping) have their own hyper-parameters which require tuning to obtain proper results. Doing so correctly is essential to ensure that our \textit{learned optimizer} provide a meaningful improvement in minimum discovery that cannot be explained by improved tuning.

\subsection{Tuning Adam and FIRE}
In our paper, all baseline results arise from a two-stage process. First, we start with a grid search: utilizing 3 variants of the learning rate for each model, with values of \{0.01, 0.005, 0.001\}. Early exploration also modified the $\beta$ parameters of Adam and the rate of increase/decrease of FIRE; however, both optimizers showed generic robustness to hyper-parameters other than learning rate changing. 

While these baselines were somewhat competitive, we further tuned these optimizers by \textit{learning} the values of the hyper-parameters for Adam and FIRE, following the \textit{Adam4p} strategy by \citet{Metz_Using}. This strategy, similar in spirit to the \textit{learned optimizers} presented in the paper, uses meta-training to update the scalar parameters that define Adam and FIRE. 3 runs were conducted, initializing with each of the learning rates in the grid search. Meta-optimization was conducted for 100 outer-steps, applied with Adam with a learning rate of $10^{-2}$. The best hyper-parameters out of the grid search and the final \textit{learned} parameters were then used to provide evaluation results.

\subsection{Tuning Basin Hopping}
Recall, Basin Hopping first uses standard optimization technique such as Adam or FIRE to optimize a network for a short number of steps. Parameters are perturbed from the minimum found, and optimization is performed once again to find a new minimum. If the new minimum is an improvement over the previous, then the new state is accepted; otherwise, the model reverts to the previous minimum.

This two-stage optimization technique has a number of hyper-parameters that require tuning. First, to simplify the setup, we fix Adam with a learning rate of $10^{-2}$ to be the standard optimizer used to descend into minima after the large perturbation steps. Experimental evidence showed that this large learning rate model allowed us to decrease the number of steps to 5000, while almost always converging to a local minima. The final parameter of significance is the size of the step taken, which is drawn from a normal distribution. Similar to the approach to tuning Adam and FIRE, we started with grid search, finding the best values out of the set \{0.2, 0.4, 0.6, 0.8\}. Additional tuning was performed by learning the size of the update step. As before, the best of these hyper-parameters was used in reporting results.

\label{app:basin}

\section{Training Details}
For the sake of reproducible results, we provide additional details about learned optimizer training, including descriptions of how tasks are selected for the meta-batches and about what it means to produce a \textit{random} initialization for particles. 

\subsection{Task Selection}
During meta-training, estimates of the meta-gradients are produced by averaging over $\sim$ 80 instantiations of atomic structure optimization problems, defined by both a random initialization (see Appendix \ref{app:harmonic}) and a corresponding empirical potential to minimize. Prior work on learned optimizers would refer to this set of problems as a task distribution and sample 80 tasks used to compute the a single meta-update. However, as the learned optimizers trained in this paper are still in the few-task regime, we instead default to sampling $\lfloor 80 / m \rfloor$ copies of all $m$ tasks.

\subsection{Harmonic Initialization}
In the context of atomic structure optimization, purely random initializations (i.e. uniform over a pre-defined box size) are problematic as atoms that are too close to one another will have very large forces early in optimization. This can result in one atom being moved far away from the others. As most molecular dynamics simulations (including ours) use a cutoff distance for atomic interactions, future optimization steps are unlikely to update and recover this atom. As more particles often corresponds to lower energies, the resultant structure will be worse off than initializations where all particles are incorporated into the final structure.

A number of strategies have been proposed to stably initialize sets of atoms or particles. One strategy is to initialize uniformly over a large box, large-enough so that the particles are unlikely to be close but small enough that cutoffs are not a problem. In practice, we found this method difficult to tune when working with a variety of systems. 

Instead, we utilize \textit{harmonic initialization}. This strategy starts by randomly initializing coordinates in a small box, of size 3.0 \AA{} for all Lennard-Jones models. Before atomic structural optimization begins, we first optimize an soft-sphere potential, which only penalizes per-particles distances when atoms are within pre-defined cutoff. Optimizing this intermediate function ensures that structural optimization does not have excessively large forces at the beginning of training. In free-space, this occurs by simply spreading particles apart.

A functional form of the soft-sphere potential is provided below:

\begin{equation}
    \sum_{i} \sum_{j > i} \begin{cases} 0 \text{ if } d_{ij} >0.1 \\
\epsilon \frac{(1 - d_{ij})^\alpha}{\alpha} \text{ otherwise}  \end{cases}
\end{equation}

where in our formulation $\epsilon = 1$ and $\alpha = 2$. By default, we use 1000 optimization steps performed by gradient descent with a learning rate of $10^{-3}$. Note, the learned optimziers are not sensitive to changes in the harmonic step count or learning rate; both defaults were chosen to allow excess time for the soft-sphere potential to be minimized to $\approx 0$.

The same strategy is used for Gupta potentials, with the only difference being that the box is increased to edge length 4.0 \AA{} to accomodate for the increased size of the atoms. For crystal optimization with the Stillinger-Weber potentials, our initialization respects the periodic boundary conditions, so particles are optimized within the pre-defined lattice.

\label{app:harmonic}

\subsection{Learned Optimizer Initialization}
All weights of the neural network used to parameterize the learned optimizer are initialized via a LeCun Normal initialization \cite{lecun2012efficient}, following the default in the FLAX library \cite{flax2020github}, except for the output layer which is initialized to have output variance of 0. This default ensures that the learned optimizer at the beginning of training does not yield divergent trajectories.

For meta-training, the parameters that have the most significant impact on performance are the $\alpha, \beta, \gamma$ used in the output parameterization. Best models used $$\alpha=0.1 \qquad \beta=1 \qquad \gamma=-3$$ We hypothesize that this setup is stable when combined with the 0 output initialization, as steps start out very small and increase in size over the course of meta-training. To large of a step early in training may yield the undesirable scenario of particles becoming too close to one another.

\begin{table*}[tb]
\caption{Transfer performance between potential types show that learned optimizers trained on Leannard-Jones can generalize to Stillinger-Weber, outperforming Adam and FIRE. However, the reverse is not true as the models trained on Silicon find very poor local minima, likely due to the removal of the periodic boundary condition.}
\label{tab:pot_transfer}
\begin{center}
\begin{small}
\begin{sc}
\begin{tabular}{l|ccc}
\toprule
&  \multicolumn{3}{c}{Mean of 150 Initializations} \\
Training Set & \shortstack[c]{Lennard-Jones \\ 13 atoms} & \shortstack[c]{Lennard-Jones \\ 75 atoms} & \shortstack[c]{SW\\ Silicon 64} \\
\midrule
\textbf{Baselines} \\
Adam & -40.6 & -380.5 & -256.8 \\
FIRE & -40.5 & -380.2 & -257.0 \\
\midrule
\textbf{Learned Optimizers} \\
Lennard-Jones \{13, 19, 31, 38 55, 75\} & -44.3 & -390.3 & -261.7 \\
Silicon 64 & -35.6 & -258.3 & -261.8 \\
\midrule
Global Minimia & -44.3 & -397.5 & -277.2 
\end{tabular}
\end{sc}
\end{small}
\end{center}
\vspace{2.0em}
\end{table*}

\subsection{Compute Costs}
Training costs vary significantly based on the system and distributed setup. For example, costs scale quadratically in the number of atoms for the Lennard-Jones and Gupta cluster and cubicly for the Stillinger-Weber models due to the three-body terms. Rough estimates of the training time are 30 GPU hours for the Lennard-Jones models (single atom type) with the main bottleneck coming from the optimization of the 75 atom system. The Gupta models took about 10 GPU hours due to the smaller size.

\section{Implementation}
As mentioned in the core body of the text, the empirical potentials make use of Jax\_MD \cite{Schoenholz_Jax} and the optimization makes use of pure JAX \cite{Bradbury_Jax}. Code will be made available shortly.

\section{Additional Results}
In this section, we present additional results that were unable to fit in the main body of the paper. These results including minimal training task distributions and results beyond atomic structure optimization further support the use of learned optimizers in these global minimization problems.

\subsection{Training with a Single Task}
The results presented in Section \ref{sec:single} show significant benefits on global minima discovery when training only on a subset of 6 optimization tasks (defined by a different number of Lennard-Jones atoms). This diverse training set provides examples of both simple `funneled' landscapes and glassy landscapes with large energy barries between minima. However, it may be possible to perform well with significantly fewer training tasks.

\begin{figure}[tb]
    \centering
    \includegraphics[width=0.46\textwidth]{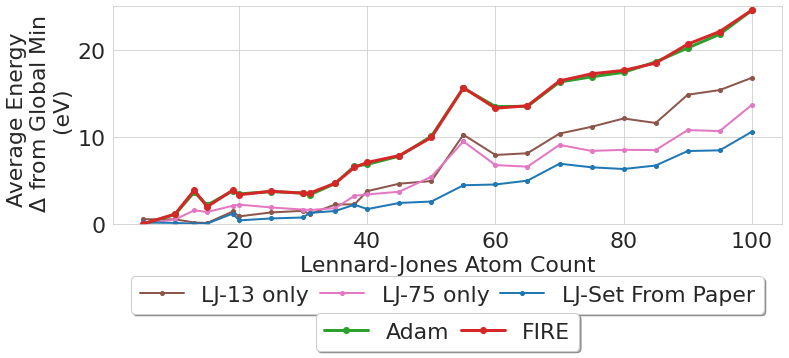}
    \vspace{-0.25em}
    \caption{Results for the Lennard-Jones single atom models, replicated with various training sets (13-atoms only, 75-atoms only, or a set of 6 different counts). Both models trained only on 1 atom count show improvements over Adam or FIRE but lack the generalization performance of training from a diverse set.}
    \vspace{-0.25em}
    \label{fig:train_single}
\end{figure}

In Figure \ref{fig:train_single}, we show a comparison between training the learned optimizer on only Lennard-Jones with 13 atoms (LJ-13), Lennard-Jones with 75 atoms (LJ-75), and the diverse Lennard-Jones set from the main body of the paper \{13, 19, 31, 38, 55, 75\}. Interestingly, the models trained only on 13 or 75 atoms often perform better than Adam and FIRE and generalize significantly beyond the respective training distributions. The 13 atom model is perhaps the most impressive as it is the cheapest to train (due to the quadratic slowdown with number of atoms); but the model taking into account diverse examples show greater generalization to large atom counts.

\subsection{Transfer Between Potentials}

Early results in Table \ref{tab:pot_transfer} also show that learned optimizers can transfer between empirical potentials; for example, the models trained on Lennard Jones clusters can transfer to the Stillinger-Weber potential despite differences in the periodic boundary and the addition of the angular term (but not vice versa). We believe that this form of generalization is most interesting and hope that future work explores this direction further; learned optimizers that train on empirical potential and can be applied to DFT simulation appears a promising avenue for significantly speeding up material design.

\end{document}